\documentclass[10pt,twocolumn,letterpaper]{article}

\usepackage[pagenumbers]{cvpr}   % To force page numbers, e.g. for an arXiv version

\usepackage[accsupp]{axessibility}
\usepackage{amsmath}
\usepackage{amssymb}
\usepackage{booktabs}
\usepackage[misc]{ifsym}
\usepackage{graphicx}
\usepackage{multirow}
\usepackage{pifont}
\usepackage{tabularx}
\usepackage{xr}

\newcommand{\cmark}{\textcolor{green}{\ding{51}}}%
\newcommand{\xmark}{\textcolor{red}{\ding{55}}}%

\usepackage[pagebackref=true,breaklinks=true,colorlinks,citecolor=blue,linkcolor=red,bookmarks=false]{hyperref}

\usepackage[capitalize]{cleveref}
\crefname{section}{Sec.}{Secs.}
\Crefname{section}{Section}{Sections}
\Crefname{table}{Table}{Tables}
\crefname{table}{Tab.}{Tabs.}

\def\paperID{688} % *** Enter the Paper ID here
\def\confName{CVPR}
\def\confYear{2025}

\makeatletter
\begin{document}

\newcolumntype{Y}{>{\centering\arraybackslash}X}

%%%%%%%%% TITLE
\title{Generative Gaussian Splatting for Unbounded 3D City Generation}

\author{%
Haozhe Xie, Zhaoxi Chen, Fangzhou Hong, Ziwei Liu
\textsuperscript{\Letter} \\
S-Lab, Nanyang Technological University\\
{%
\tt\small \{haozhe.xie, zhaoxi001, fangzhou.hong, ziwei.liu\}@ntu.edu.sg}\\
\tt\small \url{https://haozhexie.com/project/gaussian-city}
}

\twocolumn[{%
\renewcommand\twocolumn[1][]{#1}%
\maketitle
%%%%%%%%% Teaser Image
\begin{center}
  % \vspace{-3 mm}
  \centering
  \captionsetup{type=figure}
  \includegraphics[width=\textwidth]{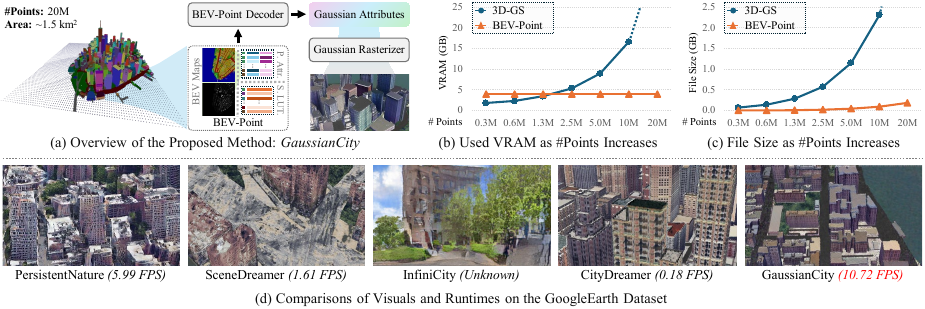}
  % \vspace{-6 mm}
  \captionof{figure}{\textbf{(a)} Benefiting from the compact BEV-Point representation, GaussianCity can generate unbounded 3D cities using 3D Gaussian splatting (3D-GS). 
  % Note that ``P.Attr'' and ``S.LUT.'' are short for ``Point Attributes'' and ``Style Look Up Table'', respectively. 
  \textbf{(b)} As the number of points increases, VRAM usage during 3D-GS training rises significantly, whereas BEV-Point, acting as a compact representation, maintains a constant VRAM usage.
  \textbf{(c)} As the number of points increases, BEV-Point exhibits significantly lower growth in file storage compared to 3D-GS.
  \textbf{(d)} The proposed \textit{GaussianCity} achieves not only superior generation quality but also the best efficiency in 3D city generation.}
  \label{fig:teaser}
\end{center}%
}]

% \renewcommand{\thefootnote}{}
% \footnotetext{\textsuperscript{\Letter} Corresponding author}

%%%%%%%%% ABSTRACT
\begin{abstract}
3D city generation with NeRF-based methods shows promising generation results but is computationally inefficient.
Recently 3D Gaussian splatting (3D-GS) has emerged as a highly efficient alternative for object-level 3D generation.
However, adapting 3D-GS from finite-scale 3D objects and humans to infinite-scale 3D cities is non-trivial. Unbounded 3D city generation entails significant storage overhead (out-of-memory issues), arising from the need to expand points to billions, often demanding hundreds of Gigabytes of VRAM for a city scene spanning 10km\textsuperscript{2}.
In this paper, we propose \textbf{GaussianCity}, a generative Gaussian splatting framework dedicated to efficiently synthesizing unbounded 3D cities with a single feed-forward pass.
Our key insights are two-fold: 
\textbf{1)} \textit{Compact 3D Scene Representation:}
We introduce BEV-Point as a highly compact intermediate representation, ensuring that the growth in VRAM usage for unbounded scenes remains constant, thus enabling unbounded city generation.
\textbf{2)} \textit{Spatial-aware Gaussian Attribute Decoder:}
We present spatial-aware BEV-Point decoder to produce 3D Gaussian attributes, which leverages Point Serializer to integrate the structural and contextual characteristics of BEV points.
Extensive experiments demonstrate that GaussianCity achieves state-of-the-art results in both drone-view and street-view 3D city generation. 
Notably, compared to CityDreamer, GaussianCity exhibits superior performance with a speedup of 60 times (10.72 FPS \textit{v.s.} 0.18 FPS).
\end{abstract}

%%%%%%%%% BODY TEXT
\section{Introduction}

The generation of 3D assets has attracted considerable attention from both academia and industry due to its significant potential applications.
With the rapid advancements in generative modeling, there have been notable achievements in 3D content generation, encompassing the production of 
objects~\cite{DBLP:conf/iclr/TangRZLZ24,DBLP:journals/ijcv/XieYZZS20,DBLP:conf/cvpr/ChenTDCFLWXWSPLL25}, 
avatars~\cite{DBLP:conf/nips/0009HMWYL23,DBLP:journals/tog/HongZPCYL22,DBLP:conf/nips/KolotourosAZBFS23}, and 
scenes~\cite{DBLP:conf/cvpr/XieCHL24,DBLP:journals/corr/abs-2404-06780,DBLP:journals/corr/abs-2501-08983}.
City generation, recognized as one of the most demanding tasks in 3D content creation, holds wide-ranging applications across domains such as gaming, animation, film, and virtual reality.

In recent years, significant progress has been made in the field of 3D city generation.
% NeRF-based methods
Both InfiniCity~\cite{DBLP:conf/iccv/LinLMCS0T23} and CityDreamer~\cite{DBLP:conf/cvpr/XieCHL24} employ NeRF~\cite{DBLP:conf/eccv/MildenhallSTBRN20} to generate unbounded photorealistic 3D cities, achieving promising results. 
However, these methods sample points at the same density and aggregate color values for all points along rays, resulting in inefficiencies during inference and loss of details.
% 3D-GS
Over the past year, 3D Gaussian splatting (3D-GS)~\cite{DBLP:journals/tog/KerblKLD23} has gained widespread use in 3D generation, offering a significantly faster rendering technique by leveraging GPU-based rasterization.
Moreover, 3D-GS provides a more flexible way to represent details, leveraging more 3D Gaussians to capture finer intricacies.
% Drawbacks of existing 3D-GS generators
However, existing 3D-GS-based generators~\cite{DBLP:conf/iclr/TangRZLZ24,DBLP:conf/cvpr/LiuZTSZLLL24,DBLP:conf/cvpr/ChungLNLL24} only produce finite-scale objects or scenes containing a limited number of 3D Gaussians.
As illustrated in Figure~\ref{fig:teaser}\textcolor{red}{(b-c)}, when the scene scales up, the demands for GPU memory (VRAM) and file storage grow dramatically, making it infeasible for unbounded scene generation.

To address these issues, we propose \textbf{GaussianCity}, the first generative Gaussian splatting for unbounded 3D city generation.
As shown in Figure~\ref{fig:teaser}\textcolor{red}{(a)}, GaussianCity leverages a highly compact scene representation, namely BEV-Point, to decouple Gaussian attributes into two major parts: position-related attributes and style-related attributes.
The position-related attributes can be further condensed into bird's-eye-view (BEV) maps, while the style-related attributes can be compressed into a style lookup table.
Only visible BEV points are considered during the rendering and optimization process, ensuring that VRAM usage is kept at a constant level.
To generate 3D Gaussian attributes from BEV points, we present BEV-Point Decoder, which leverages Point Serializer to capture structural and contextual characteristics of unstructured BEV points.
Notably, the contextual characteristics of points are inherently preserved during point sampling within a local 3D patch in NeRF, whereas these are disrupted due to the unstructured nature of BEV points.
The Gaussian Rasterizer is finally employed to render the image from the generated Gaussian attributes.

The contributions are summarized as follows:
\begin{itemize}
\item To the best of our knowledge, we propose the first 3D-GS generative model for unbounded 3D city generation with both high realism and efficiency.
\item We introduce BEV-Point as a highly compact representation, which ensures that the VRAM usage remains constant as the scene size scales up.
\item We present BEV-Point Decoder, which leverages Point Serializer to capture both structural and contextual characteristics of unstructured BEV points.
\item Extensive experiments showcase the superiority of GaussianCity over state-of-the-art methods for 3D city generation in terms of both generation quality and efficiency.
\end{itemize}

\section{Related Work}

\noindent \textbf{3D Gaussian Splatting.}
3D Gaussian splatting (3D GS)~\cite{DBLP:journals/tog/KerblKLD23} has gained considerable attention in recent months, demonstrating promising rendering results and faster performance compared to NeRF~\cite{DBLP:conf/eccv/MildenhallSTBRN20}.
% Reconstruction
Numerous methods have been proposed to integrate 3D GS into the 3D reconstruction of objects~\cite{DBLP:conf/cvpr/SzymanowiczRV24,DBLP:preprint/arxiv/2403-14621,DBLP:conf/cvpr/ZouYCLLCZ24}, avatars~\cite{DBLP:conf/cvpr/QianKSDGN24,DBLP:conf/cvpr/ShaoWLWLZFW24,DBLP:conf/cvpr/ZhengZSLZNL24}, and scenes~\cite{DBLP:conf/cvpr/LinLTLLLLWXYY24,DBLP:preprint/arxiv/2404-01133,DBLP:conf/cvpr/LuYXXWLD24}. 
Furthermore, several approaches~\cite{DBLP:conf/cvpr/LeeRSKP24,DBLP:conf/cvpr/NiedermayrSW24} have aimed to optimize the high memory consumption of 3D GS during reconstruction.
% 3D Generation
3D GS is also extensively used in the generation of 
3D objects~\cite{DBLP:preprint/arxiv/2402-05054,DBLP:conf/iclr/TangRZLZ24},
3D avatars~\cite{DBLP:conf/cvpr/AbdalYSXPKCYW24,DBLP:conf/cvpr/YuanLHMNKI24,DBLP:conf/cvpr/LiuZTSZLLL24}, and 
3D scenes~\cite{DBLP:conf/cvpr/ChungLNLL24}.
However, these methods only generate small-scale 3D assets and are not memory-efficient for representing large-scale 3D assets.
% 4D Generation
Benefiting from 3D-GS, efficiently applying it to 4D object~\cite{DBLP:preprint/arxiv/2403-12365,DBLP:preprint/arxiv/2312-17142} and human~\cite{DBLP:conf/cvpr/LeiWPLD24,DBLP:conf/cvpr/LiZWL24,DBLP:conf/cvpr/XuPLHSSBZ24} generation with deformation becomes feasible.

\noindent \textbf{Scene Generation.}
Unlike impressive models for generating objects and avatars, generating scenes presents greater challenges due to their extremely high diversity. 
% InfiniteNature, InfiniteNature-Zero
Earlier approaches~\cite{DBLP:conf/eccv/LiWSK22,DBLP:conf/iccv/LiuM0SJK21} generate scenes by synthesizing videos, which lack 3D awareness and cannot ensure 3D consistency. 
% SPADE, GANCraft
Semantic image synthesis methods~\cite{DBLP:conf/iccv/HaoMB021,DBLP:conf/cvpr/Park0WZ19,DBLP:conf/eccv/ShiSZYC22} exhibit promising results in generating scene-level content by conditioning on pixel-wise dense correspondence, such as semantic segmentation maps or depth maps.
% PersistentNature, WonderJourney, LucidDreamer
Several methods~\cite{DBLP:conf/cvpr/Chai0LIS23,DBLP:conf/cvpr/ChungLNLL24,DBLP:conf/cvpr/YuDHSRFCSSWH24} generate natural 3D scenes by performing inpainting and outpainting on RGB images or feature maps. 
However, most methods can only interpolate or extrapolate a limited distance from the input views and lack generative capabilities.
% SceneDreamer, InfiniteCity, and CityDreamer
Recent approaches~\cite{DBLP:journals/pami/ChenWL23,DBLP:conf/iccv/LinLMCS0T23,DBLP:conf/cvpr/XieCHL24,DBLP:conf/iclr/BianKXPQL25} have achieved 3D consistent scenes at an infinite scale through unbounded layout extrapolation. 
Another set of works focuses on indoor scene synthesis using costly 3D datasets~\cite{DBLP:conf/nips/0001GATTCDZGUDS22,DBLP:conf/iccv/DeVries0STS21} or CAD object  retrieval~\cite{DBLP:conf/nips/PaschalidouKSKG21}.

\section{Our Approach}

\begin{figure*}[!t]
  \resizebox{\linewidth}{!}{
    \includegraphics{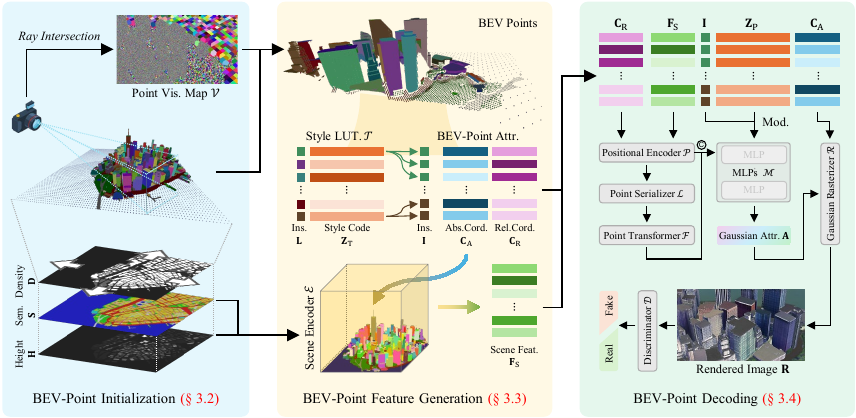}
  }
  \caption{\textbf{The framework of GaussianCity.} 
  To create an unbounded 3D city, the BEV points are firstly generated from a local patch of the BEV maps, which includes the height field $\mathbf{H}$, semantic map $\mathbf{S}$, and binary density map $\mathbf{D}$. 
  Then, the BEV-Point attributes $\left\{\mathbf{I}, \mathbf{C}_{\it A}, \mathbf{C}_{\it R}, \mathbf{F}_{\it S}\right\}$ are generated for each point and the Style Lookup Table: $\mathcal{T}(\mathbf{L}) \to \mathbf{Z}_T$ is generated for each instance. 
  Next, BEV-Point Decoder generates the Gaussian attributes $\mathbf{A}$ from BEV-Point attributes. 
  Finally, Gaussian Rasterizer $\mathcal{R}$ produces the rendered image $\mathbf{R}$.}
  \label{fig:framework}
  \vspace{-2 mm}
\end{figure*}

\subsection{Background: 3D Gaussian Splatting}

As introduced in~\cite{DBLP:journals/tog/KerblKLD23}, 3D Gaussian splatting  (3D-GS) is an explicit point-based 3D scene representation, using a set of 3D Gaussians with various attributes to model the scene.
Each Gaussian is characterized by a center $\mathbf{x} \in \mathbb{R}^3$, a scaling factor of $\mathbf{s} \in \mathbb{R}^3$, and a rotation quaternion $\mathbf{q} \in \mathbb{R}^4$.
Additionally, it maintains opacity value $\alpha \in \mathbb{R}$ and a color feature $\mathbf{c} \in \mathbb{R}^C$ for rendering, where spherical harmonics can model view-dependent effects.
These parameters can collectively be denoted by $\mathbf{A} = \left\{\mathbf{x}_i, \mathbf{s}_i, \mathbf{q}_i, \alpha_i, \mathbf{c}_i\right\}_{i=1}^N$, where $N$ is the number of 3D Gaussians.
Rendering 3D Gaussians involves projecting them onto the image plane as 2D Gaussians and performing alpha composition for each pixel in front-to-back depth order, thus determining the final color and alpha.

\subsection{BEV-Point Initialization}

Within 3D-GS, all 3D Gaussians undergo initialization with a predefined set of parameters during optimization. 
However, as the scene scales up, the VRAM usage increases dramatically, making it impractical to generate large-scale scenes.
% \zx{I'd prefer ``large-scale scenes'' instead of ``unbounded scenes'' here.}
Addressing this issue, we propose a highly compact representation namely BEV-Point.
%
% \zx{It would be great to put this paragraph before introducing details of BEV-Point. In other words, we need one / two sentences in L94 as a high-level introduction to BEV-Point, offering intuition of this representation.}
In the BEV-Point representation, only visible BEV points are retained since only they impact the appearance of the current frame. 
It ensures that VRAM usage remains constant because the number of visible BEV points does not increase with the scene scale, given fixed camera parameters. 
% \zx{More precisely, the number of visible points does not increase with scene scale given fixed camera parameters.}

In a local patch of a bird's-eye-view (BEV) map comprising a semantic map $\mathbf{S}$ and a height field $\mathbf{H}$, a collection of BEV points within this patch can be produced by extruding the pixels in the semantic map $\mathbf{S}$ according to the corresponding values in the height field $\mathbf{H}$.
% \zx{Remove: In contrast to CityDreamer~\cite{DBLP:conf/cvpr/XieCHL24}, which employs uniform density sampling along rays}, 
We further introduce the binary density map $\mathbf{D}$ to adjust the sampling density for different semantic categories.
This is driven by the observation that certain categories exhibit simpler textures (\textit{e.g.}, roads, water areas), allowing for reduced density to manage computational costs, while other categories (\textit{e.g.}, building fa\c{c}ades) possess intricate textures, necessitating a greater number of points for representation.
The coordinates of the generated BEV points, denoted as $\mathbf{C}_F \in \mathbb{R}^{N_{\rm pt} \times 3}$, can be generated as
\begin{equation}
  \mathbf{C}_F = \left\{(x, y, z)~|~\mathbf{H}_{(x, y)} \le z~{\rm and}~\mathbf{D}_{(x, y)} = 1 \right\}
\end{equation}
where $N_{\rm pt}$ is the number of BEV points.

Benefiting from the binary density map $\mathbf{D}$, a significant number of redundant BEV points have been omitted.
However, the remaining number, typically 20 million BEV points, is still too large for optimization.
To address this issue, we additionally conduct ray intersection to obtain the binary visibility map 
$\mathcal{V}: (x, y, z) \to v$, where $v \in \left\{0, 1\right\}$, for filtering out the visible BEV points.
Therefore, the coordinates of the visible BEV points $\mathbf{C}_A \in \mathbb{R}^{N_{\rm pt} \times 3}$ can be generated as
\begin{equation}
  \mathbf{C}_A = \left\{(x, y, z)~|~(x, y, z) \in \mathbf{C}_F~{\rm and}~\mathcal{V}(x, y, z) = 1 \right\}
\end{equation}

\subsection{BEV-Point Feature Generation}

The features in the BEV-Point representation can be divided into three categories: instance attributes, BEV-Point attributes, and style look-up table. 
The instance attributes encompass fundamental details such as Instance label, size, and center coordinates for each instance.
The BEV-Point attributes determine the appearance within the instance,
while the style look-up table controls the style variation across instances.

\noindent \textbf{Instance Attributes.}
The semantic map $\mathbf{S}$ provides semantic labels for BEV points.
Following CityDreamer~\cite{DBLP:conf/cvpr/XieCHL24}, the instance map $\mathbf{Q}$ is introduced to handle the diversity of buildings and cars in urban environments.
Specifically,
\begin{equation}
  \mathbf{Q} = {\rm Inst}(\mathbf{S})
\end{equation}
where $\rm{Inst}(\cdot)$ denotes instantiation on semantic maps by detecting connected components.
Therefore, the instance label $\mathbf{I} \in \mathbb{R}^{N_{\rm pt} \times 1}$ for BEV points can be computed as
\begin{equation}
  \mathbf{I} = \left\{ \mathbf{Q}_{(x, y)}~|~(x, y, z) \in \mathbf{C}_A \right\}
\end{equation}
Obviously, $\mathbf{Q}_{(x, y)} \in \mathbf{L}$,
where $\mathbf{L} = \left\{1, 2, \dots, N_{\rm ins}\right\}$ and $N_{\rm ins}$ is the number of instances.
The size $\mathcal{S}(l) \in \mathbb{R}^3$ represents the size of the 3D bounding box of the instance $l$, where $l \in \mathbf{L}$.
The center $\mathcal{C}(l) \in \mathbb{R}^3$ denotes the coordinates of the center of the bounding box of the instance $l$.

\noindent \textbf{BEV-Point Attributes.}
%
% \zx{``absolute'' coordinate and ``relative'' coordinate are confused. It would be great to specify the coordinate system.}
In BEV-Point initialization, the absolute coordinate $\mathbf{C}_A$ is generated, with the origin set at the center of the world coordinate system.
Besides the absolute coordinate, the relative coordinate $\mathbf{C}_R \in \mathbb{R}^{N_{\rm pt} \times 3}$ is introduced, with its origin set at the center of each instance, to specify the normalized point coordinates relative to the instance.
Specifically,
\begin{equation}
  \mathbf{C}_R = \left\{\mathbf{C}_R^i~|~\mathbf{C}_R^i \in \mathbb{R}^3 \right\}_{i=1}^{N_{\rm pt}}
\end{equation}
where $\mathbf{C}_R^i$ can be derived as
\begin{equation}
  \mathbf{C}_R^i = \frac{2\left(\mathbf{C}_A^i - \mathcal{C}(\mathbf{I}^i)\right)}{\mathcal{S}(\mathbf{I}^i)}
\end{equation}
where $\mathbf{C}_A^i$ and $\mathbf{I}^i$ denote the $i$-th values in $\mathbf{C}_A$ and $\mathbf{I}$, respectively.

During generation, integrating contextual information for BEV points becomes essential. 
It is achieved by introducing scene features $\mathbf{F}_S \in \mathbb{R}^{N_{\rm pt} \times C_{F_S}}$ derived from BEV maps and indexed using absolute coordinates $\mathbf{C}_A$.
Specifically,
\begin{equation}
  \mathbf{G} = \mathcal{E}(\mathbf{H}, \mathbf{S})
\end{equation}
\begin{equation}
  \mathbf{F}_S = \left\{\mathbf{G}_{(x, y)}~|~(x, y, z) \in \mathbf{C}_A \right\}
\end{equation}
where $\mathcal{E}$ denotes the Scene Encoder.
$\mathbf{G}$ is with the same size of $\mathbf{H}$ and $\mathbf{S}$.

\noindent \textbf{Style Look-up Table.}
In 3D-GS, the appearance of 3D Gaussians is defined by the attributes of each Gaussian.
As the number of 3D Gaussians increases, the demands on VRAM and file storage grow significantly, making unbounded scene generation infeasible.
To further reduce computational costs, the appearances of instances are encoded into a set of latent vectors $\mathbf{Z}_T \in \mathbb{R}^{N_{\rm ins} \times C_Z}$, \textit{i.e.},
\begin{equation}
  \mathbf{Z}_T = \left\{\mathbf{z}_T^i | \mathbf{z}_T^i \in \mathbb{R}^{C_Z}  \right\}_{i = 1}^{N_{\rm ins}}
\end{equation}
The style look-up table $\mathcal{T}$ queries the style code $\mathbf{z}_T^i \sim \mathcal{N}(0, 1)$ for the instance $l$. 
Specifically,
\begin{equation}
  \mathbf{z}_T^i = \mathcal{T}(l)
\end{equation}

\subsection{BEV-Point Decoding}

The BEV-Point decoder is designed to generate the Gaussian Attributes $\mathbf{A}$ using the BEV-Point features. 
It comprises five key modules: positional encoder, point serializer, point transformer, modulated MLP, and Gaussian rasterizer.

\noindent \textbf{Positional Encoder.}
Rather than feed coordinates directly into the subsequent networks, the positional encoder transforms each point coordinate and corresponding features into a higher-dimensional positional embedding $\mathbf{F}_P \in \mathbb{R}^{N_{\rm pt} \times C_{F_P}}$ as follows
\begin{equation}
  \mathbf{F}_P = \mathcal{P}\left({\rm Concat}(\mathbf{C}_R, \mathbf{F}_S)\right)
\end{equation}
where $\mathcal{P}(\cdot)$ is the positional encoding function that is applied individually to each value in the given feature $x$.
\begin{equation}
  \mathcal{P}(x) = \left\{ \sin(2^i\pi x), \cos(2^i\pi x) \right\}_{i=0}^{N_{\rm PE} - 1}
\end{equation}

\noindent \textbf{Point Serializer.}
Unlike NeRF~\cite{DBLP:conf/eccv/MildenhallSTBRN20}, which maintains spatial correlation among sampled points along rays, BEV points and 3D Gaussians are unstructured and unordered due to their irregular nature of point clouds.
Therefore, directly applying Multi-layer Perceptrons (MLPs) to generate Gaussian attributes from 
$\mathbf{F}_P$ may not yield optimal results, as MLPs do not fully consider the structural and contextual characteristics of point clouds.
%
% In point cloud processing, approaches KNN-based~\cite{DBLP:conf/nips/QiYSG17} and voxelization-based~\cite{DBLP:conf/eccv/XieYZMZS20} methods have been introduced to incorporate context information.
% However, KNN-based methods suffer from computational inefficiencies, whereas voxelization-based methods are inefficient in terms of VRAM usage.

To transform unstructured BEV points into a structured format, we present the point serializer $\mathcal{L}: (x, y, z) \to o$, where $o \in \mathbb{Z}$, to convert the point coordinates into an integer reflecting its order within the given BEV points.
\begin{equation}
  \mathcal{L}(x, y, z, g) = \left\lfloor \frac{x}{g^2} + \frac{y}{g} + z \right\rfloor
  \label{eq:serialization}
\end{equation}
where $g$ is the grid size that controls the scale of the discrete space for serialization.
This ordering ensures that points are rearranged based on the spatial structure defined by the point serializer, resulting in neighboring points in the data structure being close together in space.
Thus, the serialized feature $\mathbf{F}_P^S \in \mathbb{R}^{N_{\rm pt} \times C_{F_P}}$ can be derived as
\begin{equation}
  \mathbf{F}_P^S = \left\{\mathbf{F}_P^{O_i}\right\}_{i = 1}^{N_{\rm pt}}
\end{equation}
where $\mathbf{F}_P^{O_i}$ denotes the features of the ${O_i}$-th BEV point in $\mathbf{F}_P$.
${O_i}$ denotes the order generated by the point serializer.

\noindent \textbf{Point Transformer.}
After serialization, the features of BEV points can be processed by a modern Transformer $\mathcal{F}$~\cite{DBLP:conf/cvpr/WuJWLLQOHZ24}, producing the feature $\mathbf{F}_T \in \mathbb{R}^{N_{\rm pt} \times C_{F_T}}$.
\begin{equation}
  \mathbf{F}_T = \mathcal{F}(\mathbf{F}_P^S)
\end{equation}

\noindent \textbf{Modulated MLPs.}
%
% \zx{MLPs do not characterize any unique process here. Modulation and how to interact with style code and instance labels are of more importance.}
The attributes of 3D Gaussians $\mathbf{A} \in \mathbb{R}^{N_{\rm pt} \times C_A}$ are generated by applying MLPs $\mathcal{M}$~\cite{DBLP:conf/cvpr/KarrasLAHLA20}, which modulate the style code $\mathbf{Z}_P$ and instance labels $\mathbf{I}$ with the features of BEV points.
\begin{equation}
  \mathbf{A} = \mathcal{M}({\rm Concat}(\mathbf{F}_P, \mathbf{F}_T), \mathbf{Z}_P, \mathbf{I})
\end{equation}
where $\mathbf{Z}_P = \left\{\mathcal{T}(\mathbf{I}^i) \right\}_{i = 1}^{N_{\rm pt}}$.

\noindent \textbf{Gaussian Rasterizer.}
Given the camera intrinsic and extrinsic parameters $\mathbf{K} \in \mathbb{R}^{3 \times 3}$ and $\mathbf{T} \in \mathbb{R}^{4 \times 4}$, the image can be rendered with the Gaussian rasterizer $\mathcal{R}$.
\begin{equation}
  \mathbf{\hat{R}} = \mathcal{R}(\mathbf{C}_A, \mathbf{A}, \mathbf{K}, \mathbf{T})
\end{equation}
During rasterization, default values are employed if the required attributes are not generated in $\mathbf{A}$.
These defaults include scale factor $\mathbf{s} = \mathbf{1}$, rotation quaternion $\mathbf{q} = [1, 0, 0, 0]$, and opacity $\alpha = 1$.

\subsection{Loss Functions}
The generator is trained using a hybrid objective that combines reconstruction loss with adversarial learning loss.
Specifically, we leverage the L1 loss, VGG loss~\cite{DBLP:conf/eccv/JohnsonAF16}, and GAN loss~\cite{DBLP:preprint/arxiv/1705.02894} in this combination.
\begin{equation}
  \ell = \lambda_{\rm L1} \lvert\lvert \mathbf{\hat{R}} - \mathbf{R} \lvert\lvert
       + \lambda_{\rm VGG} {\rm VGG}(\mathbf{\hat{R}}, \mathbf{R})
       + \lambda_{\rm GAN} {\rm GAN}(\mathbf{\hat{R}}, \mathbf{S}_G)
\end{equation}
where $\mathbf{R}$ denotes the ground truth image.
$\mathbf{S}_G$ is the semantic map in perspective view.

\section{Experiments}

\begin{table*}[!t]
  \setlength\extrarowheight{1pt}
  \caption{\textbf{Quantitative comparison on GoogleEarth and KITTI-360.} The best results are highlighted in bold. Note that ``RT.'' denotes ``Runtime'', measured in milliseconds on an NVIDIA Tesla A100.}
  \label{tab:quatitative-comp}
  % \vspace{-2 mm}
  \begin{tabularx}{\linewidth}{lYYYYY|lYY}
    \toprule
      \multicolumn{6}{c|}{GoogleEarth} & 
      \multicolumn{3}{c}{KITTI-360} \\
      \cline{1-6} \cline{7-9}
      Method  & FID $\downarrow$ & KID $\downarrow$ 
              & DE $\downarrow$  & CE $\downarrow$ 
              & RT. $\downarrow$ &
      Method  & FID $\downarrow$ & KID $\downarrow$ \\
    \midrule
      SGAM~\cite{DBLP:conf/nips/ShenMW22}               & 
      277.64     & 0.358      & 0.575      & 239.291    & 193      &
      StyleGAN~\cite{DBLP:conf/cvpr/KarrasLAHLA20}      &
      31.9       & 0.021 \\
      Pers.Nature~\cite{DBLP:conf/cvpr/Chai0LIS23}      & 
      123.83     & 0.109      & 0.326      & 86.371     & 167      &
      GSN~\cite{DBLP:conf/iccv/DeVries0STS21}           &
      160.0      & 0.114 \\
      SceneDreamer~\cite{DBLP:journals/pami/ChenWL23}   &
      213.56     & 0.216      & 0.152      & 0.186      & 620      &
      GIRAFFE~\cite{DBLP:conf/cvpr/Niemeyer021}         &
      112.1      & 0.117 \\
      CityDreamer~\cite{DBLP:conf/cvpr/XieCHL24}        &
      97.38      & 0.096      & 0.147      & 0.060      & 5580     &
      UrbanGIR.~\cite{DBLP:conf/iccv/YangYGX0L23}       &
      39.6       & 0.036 \\
      GaussianCity                                      & 
      \bf{86.94} & \bf{0.090} & \bf{0.136} & \bf{0.057} & \bf{93}  & 
      GaussianCity                                      & 
      \bf{29.5}  & \bf{0.017} \\
    \bottomrule
  \end{tabularx}
\end{table*}

\subsection{Datasets}

\noindent \textbf{GoogleEarth}~\cite{DBLP:conf/cvpr/XieCHL24}\textbf{.}
The GoogleEarth dataset, sourced from Google Earth Studio, comprises 400 orbit trajectories captured over Manhattan and Brooklyn. 
Each trajectory comprises 60 images, with orbit radiuses spanning from 125 to 813 meters and altitudes ranging from 112 to 884 meters. 
In addition to the images, GoogleEarth provides camera intrinsic and extrinsic parameters, along with 3D semantic and building instance segmentation.

\noindent \textbf{KITTI-360}~\cite{DBLP:journals/pami/LiaoXG23}\textbf{.}
KITTI-360 is an extensive outdoor sub-urban dataset known for its intricate scene geometry.
Within this dataset, scenes are rich in highlights and shadows, resulting in significant variations in the appearance of objects across different scenes.
Captured in urban scenarios spanning approximately 73.7km of driving distance, KITTI-360 provides comprehensive 3D bounding box annotations for various classes such as buildings, cars, roads, vegetation, and more.

\subsection{Evaluation Protocols}

\noindent \textbf{FID and KID.}
Fr\'echet Inception Distance (FID)~\cite{DBLP:conf/nips/HeuselRUNH17} and Kernel Inception Distance (KID)~\cite{DBLP:conf/iclr/BinkowskiSAG18} are metrics for the quality of generated images.
Following CityDreamer~\cite{DBLP:conf/cvpr/XieCHL24}, FID and KID are calculated based on 15K generated frames on the GoogleEarth dataset. 
As for the KITTI-360 dataset, FID and KID are computed using 5K generated frames, consistent with the settings of UrbanGIRAFFE~\cite{DBLP:conf/iccv/YangYGX0L23}.

\noindent \textbf{Depth Error.}
We employ depth error (DE) to assess the 3D geometry following EG3D~\cite{DBLP:conf/cvpr/ChanLCNPMGGTKKW22}. 
The pseudo ground truth depth maps for the generated frames are obtained from a pretrained model~\cite{DBLP:journals/pami/RanftlLHSK22}. 
For NeRF-based methods, the depth maps are generated by the accumulation of density $\sigma$. 
Conversely, GaussianCity allows for the direct acquisition of depth maps by rasterization.
DE is calculated as the L2 distance between the two normalized depth maps.
We evaluate DE using 100 frames for each assessed method following CityDreamer~\cite{DBLP:conf/cvpr/XieCHL24}.

\noindent \textbf{Camera Error.}
Following CityDreamer~\cite{DBLP:conf/cvpr/XieCHL24}, the camera error (CE) is introduced to evaluate the multi-view consistency.
CE quantifies the disparity between the inference camera trajectory and the estimated camera trajectory from COLMAP~\cite{DBLP:conf/cvpr/SchonbergerF16},
computed as the scale-invariant normalized L2 distance between the reconstructed and generated camera poses.

\noindent \textbf{Runtime.}
Runtime is used to gauge efficiency. It measures the time taken to generate a 960x540 image on the GoogleEarth dataset.
All timings are conducted on an NVIDIA Tesla A100, excluding IO time, and averaged over 100 iterations.

\subsection{Implementation Details}

\noindent \textbf{Hyperparameters.}
The grid size $g$ is set to $0.01$ and $N_{\rm PE}$ is set to $64$.
The feature channels $C_Z$, $C_{F_S}$, $C_{F_P}$, and $C_{F_T}$ are set to $256$, $61$, $1280$, and $256$, respectively.
The Gaussian attributes $\mathbf{A}$ consist of RGB values only (\textit{i.e.}, $C_A = 3$).
Initially, the loss weights $\lambda_{\rm L1}$, $\lambda_{\rm VGG}$, and $\lambda_{\rm GAN}$ are set to 10, 10, and 0.5, respectively. 
Afterward, $\lambda_{\rm L1}$ and $\lambda_{\rm VGG}$ are reduced by half every 20,000 epochs.

\noindent \textbf{Training Details.}
The generator is trained using an Adam optimizer with $\beta = (0, 0.999)$ and a learning rate of $10^{-4}$. 
The discriminator is optimized using an Adam optimizer with $\beta = (0, 0.999)$ and a learning rate of $5 \times 10^{-7}$.
Training continues for 200,000 iterations with a batch size of 8 on eight NVIDIA Tesla V100 GPUs. 
The images are randomly cropped to sizes of 448$\times$448 for GoogleEarth and 448$\times$224 for KITTI-360.

\begin{figure*}[!t]
  \resizebox{\linewidth}{!}{
    \includegraphics{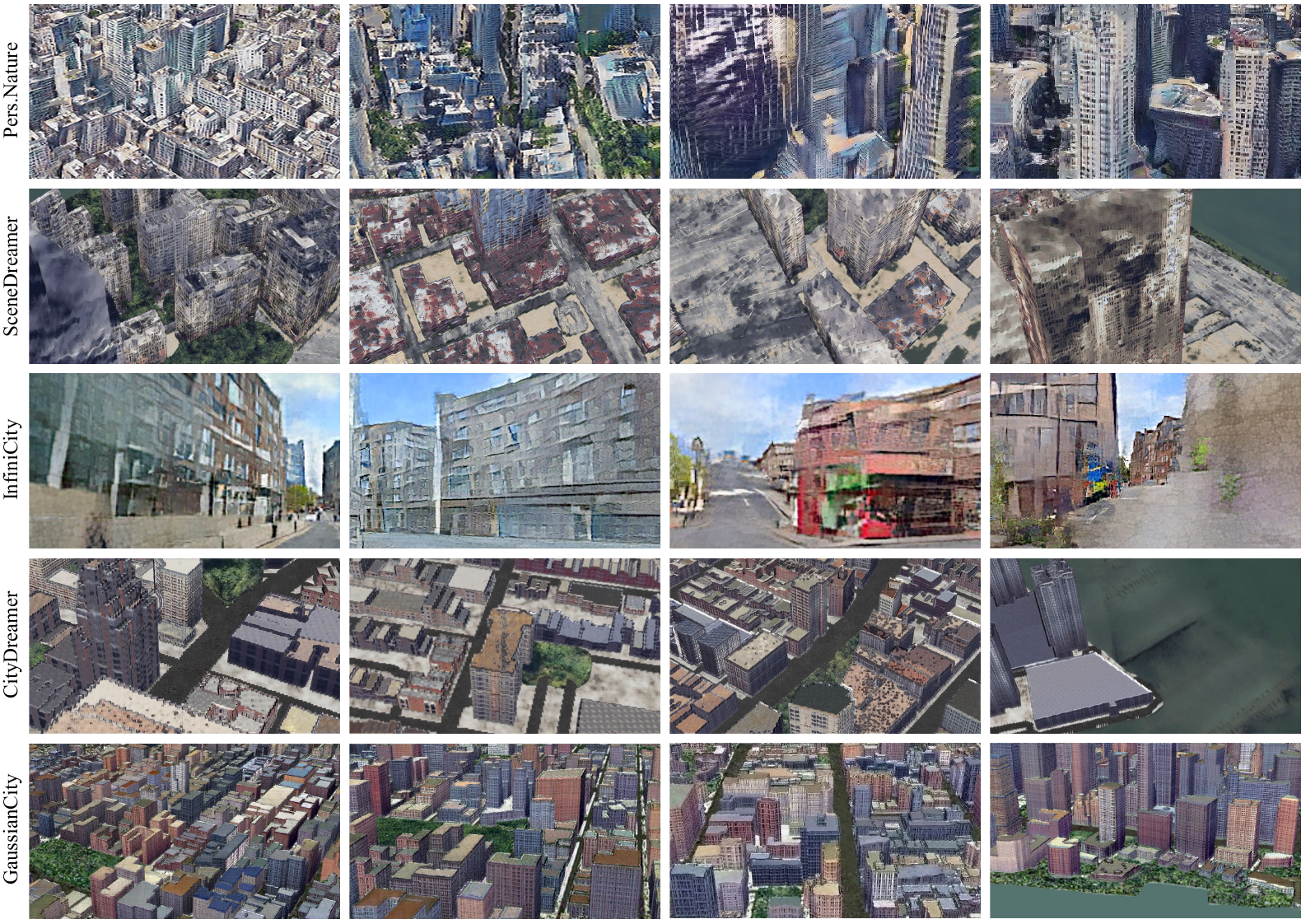}
  }
  \caption{\textbf{Qualitative comparison on GoogleEarth.} Note that ``Pers.Nature'' is short for PersistentNature~\cite{DBLP:conf/cvpr/Chai0LIS23}. The visual results of InfiniCity~\cite{DBLP:conf/iccv/LinLMCS0T23} are provided by the authors since the source code is not accessible.}
  \label{fig:qualitative-google-earth}
  \vspace{-4.5 mm}
\end{figure*}

\subsection{Main Results}

\noindent \textbf{Comparison Methods.}
On the GoogleEarth dataset, we compare the proposed method to SGAM~\cite{DBLP:conf/nips/ShenMW22}, PersistentNature~\cite{DBLP:conf/cvpr/Chai0LIS23}, SceneDreamer~\cite{DBLP:journals/pami/ChenWL23}, InfiniCity~\cite{DBLP:conf/iccv/LinLMCS0T23}, and CityDreamer~\cite{DBLP:conf/cvpr/XieCHL24}, following the protocol established by CityDreamer. 
Except for InfiniCity, whose code is not accessible, we retrain the remaining methods using the released code on the GoogleEarth dataset to ensure fair comparisons. 
On the KITTI-360 dataset, we compare our method to StyleGAN2~\cite{DBLP:conf/cvpr/KarrasLAHLA20}, GSN~\cite{DBLP:conf/iccv/DeVries0STS21}, GIRAFFE~\cite{DBLP:conf/cvpr/Niemeyer021}, and UrbanGIRAFFE~\cite{DBLP:conf/iccv/YangYGX0L23}, following the protocol outlined by UrbanGIRAFFE. 
As the training code and pretrained model for UrbanGIRAFFE are unavailable, the results of UrbanGIRAFFE are directly obtained from \cite{DBLP:conf/iccv/YangYGX0L23}.

\begin{figure*}[!t]
  \resizebox{\linewidth}{!}{
    \includegraphics{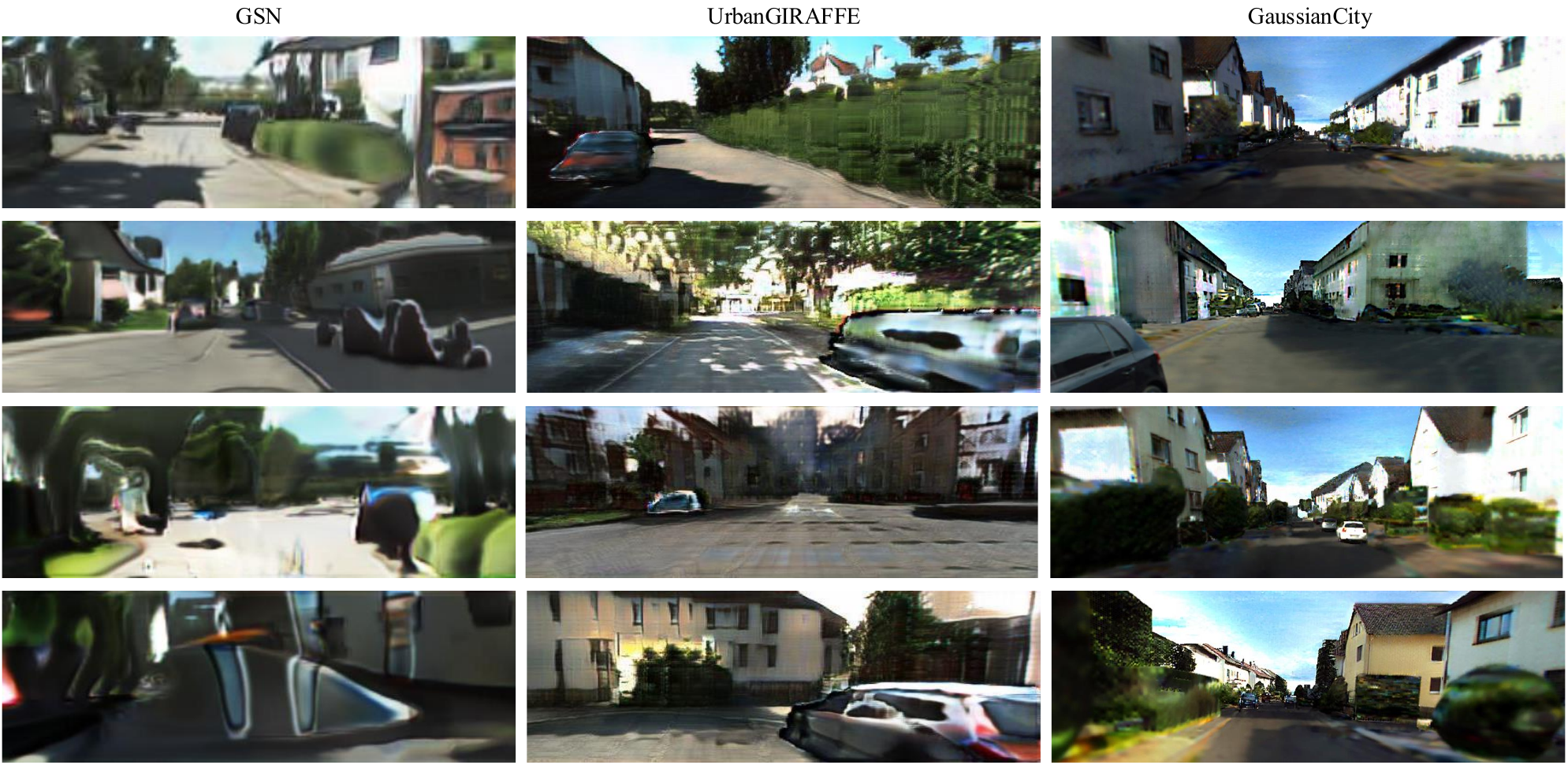}
  }
  \caption{\textbf{Qualitative comparison on KITTI-360.} The visual results of UrbanGIRAFFE~\cite{DBLP:conf/iccv/YangYGX0L23} are provided by the authors since the training code and pretrained model are unavailable.}
  \label{fig:qualitative-kitti360}
\end{figure*}

\begin{figure*}
  \centering
  \resizebox{\linewidth}{!}{
    \includegraphics{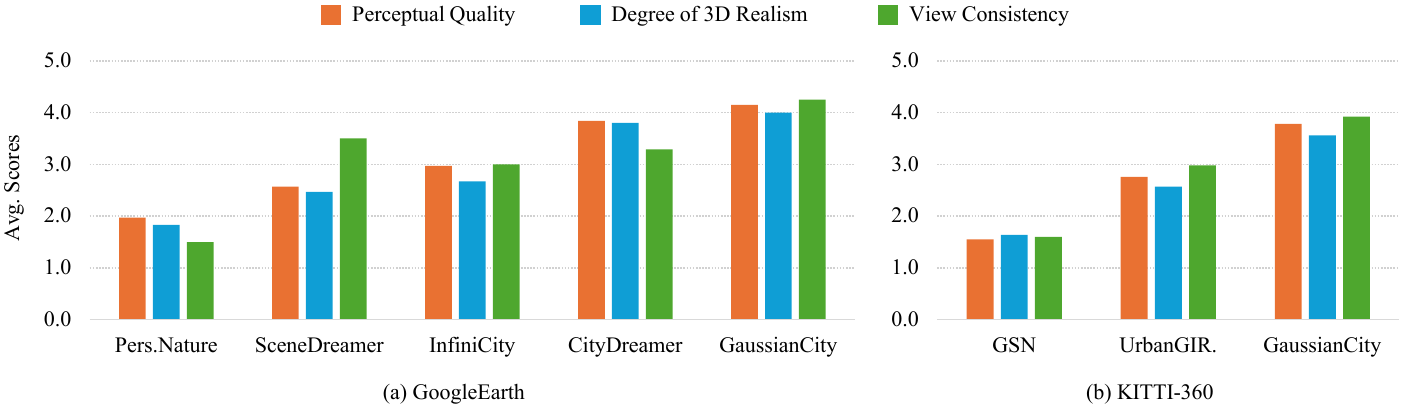}
  }
  \caption{\textbf{User study on GoogleEarth and KITTI-360.} All scores are in the range of 5, with 5 indicating the best. Note that ``Pers.Nature'' and ``UrbanGIR.'' denotes PersistentNature~\cite{DBLP:conf/cvpr/Chai0LIS23} and UrbanGIRAFFE~\cite{DBLP:conf/iccv/YangYGX0L23}, respectively.}
  \label{fig:user-study}
  \vspace{-2 mm}
\end{figure*}

\noindent \textbf{Comparison on GoogleEarth.}
Figure~\ref{fig:qualitative-google-earth} shows qualitative comparisons against baselines on the GoogleEarth dataset.
PersistentNature utilizes a tri-plane representation, which faces difficulties in generating realistic renderings.
Both SceneDreamer and InfiniCity employ voxel grids as their representation, yet they still encounter significant structural distortions in buildings due to all buildings being assigned the same semantic label.
CityDreamer and GaussianCity both introduce instance labels, achieving improved generation results. 
However, CityDreamer composites two neural fields, resulting in artifacts at the seams of images. 
In contrast, GaussianCity achieves better visual results and significantly lower runtime, as indicated in Table~\ref{tab:quatitative-comp}.

\noindent \textbf{Comparison on KITTI-360.}
Figure~\ref{fig:qualitative-kitti360} illustrates qualitative comparisons against baseline methods on the KITTI-360 dataset.
GSN uses an implicit representation named latent grid, which can easily introduce structural distortion in street-view generation.
UrbanGIRAFFE employs a voxel grid as its 3D representation, but due to the resolution limitations of the voxel grid, the generated scenes may exhibit jagged artifacts.
In contrast, GaussianCity adopts a more flexible point cloud as its representation, thus yielding superior visual results. 
The lower FID and KID scores in Table~\ref{tab:quatitative-comp} also attest to the effectiveness of the proposed method.

\noindent \textbf{User Study.}
To more accurately evaluate the 3D consistency and quality of the unbounded 3D city generation, we conduct an output evaluation~\cite{DBLP:journals/ftcgv/BylinskiiHHHZ23}, following the user study conducted in CityDreamer.
In this study, we requested feedback from 20 volunteers to assess each generated camera trajectory based on three criteria: 1) the perceptual quality of the imagery, 2) the level of 3D realism, and 3) the 3D view consistency. 
Ratings are provided on a scale of 1 to 5, with 5 indicating the highest rating. 
The results are shown in Figure~\ref{fig:user-study}, demonstrating that the proposed method significantly surpasses the baselines by a considerable margin.

\subsection{Ablation Studies}

\begin{table}[!t]
  \caption{\textbf{Effectiveness of BEV-Point Representation.} Note that ``RayInt.'' denotes ``Ray Intersection'', respectively. The VRAM usage and inference time are measured on an NVIDIA Tesla A100.}
  \label{tab:ablation-bev-point}
  \vspace{-2 mm}
  \begin{tabularx}{\linewidth}{cc|ccc}
    \toprule
      Visibility & Density  & \#Points   & VRAM (G) & Time (s) \\
    \midrule
      \xmark     & \xmark   & 31M        & \multicolumn{2}{c}{Out of Memory} \\
      \xmark     & \cmark   & 20M        & \multicolumn{2}{c}{Out of Memory} \\
    \midrule
      Region     & \cmark   & 1.4M       & \multicolumn{2}{c}{Out of Memory} \\
      Instance   & \cmark   & 680K       & 10.34     & 0.16 \\
    \midrule
      RayInt.    & \xmark   & 38.0K      & 1.41      & 0.10 \\
      RayInt.    & \cmark   & \bf{31.7K} & \bf{1.39} & \bf{0.09} \\
    \bottomrule
  \end{tabularx}
\end{table}

\begin{table}[!t]
  \caption{\textbf{Effectiveness of BEV-Point Decoder.} Note that ``Ser.'' and ``Pt.Trans.'' denote ``Point Serializer'' and ``Point Transformer'', respectively.}
  \label{tab:ablation-bev-point-decoder}
  \vspace{-2 mm}
  \begin{tabularx}{\linewidth}{Yc|YYYY}
    \toprule
      Ser.             & Pt.Trans.        & 
      FID $\downarrow$ & KID $\downarrow$ & 
      DE $\downarrow$  & CE $\downarrow$ \\
    \midrule
      \xmark           & \xmark     &
      151.27           & 0.179      &
      0.185            & 0.135 \\
      \xmark           & \cmark     &
      119.40           & 0.138      &
      0.159            & 0.118 \\
      \cmark           & \cmark     & 
      \bf{86.94}       & \bf{0.090} & 
      \bf{0.136}       & \bf{0.057} \\
    \bottomrule
  \end{tabularx}
\end{table}

\begin{table}[!t]
  \caption{\textbf{Comparison of different serialization methods.} Note that ``Eq.~\ref{eq:serialization}'' denote the serialization used in Point Serializer.}
  \label{tab:ablation-serialization}
  \vspace{-2 mm}
  \begin{tabularx}{\linewidth}{YY|YYYY}
    \toprule
      Eq.~\ref{eq:serialization} & 
      Hilbert          & 
      FID $\downarrow$ & KID $\downarrow$ & 
      DE $\downarrow$  & CE $\downarrow$ \\
    \midrule
      \cmark           & \xmark     & 
      86.94            & 0.090      & 
      0.136            & 0.057           \\
      \xmark           & \cmark     & 
      86.72            & 0.088      & 
      0.136            & 0.056      \\
      \cmark           & \cmark     & 
      \bf{86.28}       & \bf{0.083} & 
      \bf{0.135}       & \bf{0.055} \\
    \bottomrule
  \end{tabularx}
  \vspace{-4 mm}
\end{table}

\noindent \textbf{Effectiveness of BEV-Point Representation.}
The proposed BEV-Points, a compact scene representation, are essential for enabling 3D-GS to generate unbounded scenes by 
\textbf{1)} considering point visibility during generation and 
\textbf{2)} adjusting sampling density for different semantic categories using a binary density map.
Table~\ref{tab:ablation-bev-point} compares different methods for considering point visibility, including no consideration, region-based, instance-based, and ray-intersection-based strategies for point selection.
All methods keep the number of points constant: ``region-based'' selects points within the camera coverage, ``Instance-based'' selects visible instances, and ``ray-intersection-based'' identifies visible points through ray intersection~\cite{DBLP:conf/eurographics/AmanatidesW87}.
The "ray-intersection-based" method is the most efficient compared to "region-based" and "instance-based" approaches, eliminating the most invisible points and reducing the total number of points by 97.7\% and 96\%, respectively.
The density map is crucial in the BEV-Point representation because it reduces the number of points for semantic categories that have less texture and need fewer points.
As shown in Table~\ref{tab:ablation-bev-point}, removing the density map increases the point count by 20\%-50\%.
As shown in Figure~\ref{fig:teaser}\textcolor{red}{(b-c)}, the proposed BEV-Point representation maintains constant VRAM usage during training, unlike the original 3D-GS~\cite{DBLP:journals/tog/KerblKLD23}, which is impractical for unbounded scene generation due to excessive VRAM consumption.
Additionally, it significantly reduces file storage as the number of points increases.

\noindent \textbf{Effectiveness of BEV-Point Decoder.}
BEV-Point Decoder consists of two key components: \textit{Point Serializer} and \textit{Point Transformer}.
As shown in Table~\ref{tab:ablation-bev-point-decoder} and Figure~\ref{fig:ablation-bev-point-decoder}, removing both components leads to a significant degeneration due to the absence of spatial correlations.
Introducing only the Point Transformer partially establishes spatial correlation, but it does not achieve the same effectiveness as using both components together.

\noindent \textbf{Effectiveness of Serialization Methods.}
%
% Point serializer employs Equation~\ref{eq:serialization} to structure the unordered BEV points. 
To investigate the impact of different serialization methods in point serializer, we compare the effects of using Equation~\ref{eq:serialization}, Hilbert curve~\cite{DBLP:journals/ma/Hilbert91} adopted in~\cite{DBLP:conf/eccv/ChenZCY22,DBLP:conf/cvpr/WuJWLLQOHZ24}, and both simultaneously on the generation results.
As shown in Table~\ref{tab:ablation-serialization}, different serialization methods have minimal impact on the final generation results. 
We employ Equation~\ref{eq:serialization} in Point Serializer because of its reduced computational complexity.
Kindly refer to Figure~\ref{fig:ablation-serialization} for more qualitative comparisons.

\subsection{Limitations}
While our method demonstrates promising results for unconditional 3D city generation, it still has several limitations. 
Firstly, the BEV-Point Initialization elevates points to a height based on the corresponding value in the height field, which assumes buildings adhere to the Manhattan assumption and cannot represent hollow structures. 
Secondly, the BEV-Point decoder does not fully exploit the expressive capacity of 3D-GS, as it only predicts RGB in the Gaussian attributes.
However, predicting an excessive number of attributes simultaneously may destabilize the training process.
Kindly refer to the discussion in Section~\ref{sec:gaussian-attributes}.

\section{Conclusion}

In this paper, we introduce GaussianCity for generating unbounded 3D cities.
It is the first method to employ 3D-GS for unbounded scene generation. 
We introduce BEV-Point, a highly compact scene representation to mitigate the substantial increase in VRAM usage associated with representing large-scale scenes using 3D-GS.
Additionally, we present BEV-Point Decoder, which leverages Point Serializer to capture the structural and contextual characteristics of unordered BEV points.
Compared to previous methods, GaussianCity achieves better visual results and a 60-fold increase in runtime speed compared to CityDreamer.
Extensive experimental results demonstrate the effectiveness of the proposed GaussianCity in generating both drone-view and street-view in city scenes.

\noindent \textbf{Acknowledgements.}
% \section*{Acknowledgements and Disclosure of Funding}
%
This study is supported by the Ministry of Education, Singapore, under its MOE AcRF Tier 2 (MOET2EP20221- 0012, MOE-T2EP20223-0002), NTU NAP, and under the RIE2020 Industry Alignment Fund – Industry Collaboration Projects (IAF-ICP) Funding Initiative, as well as cash and in-kind contribution from the industry partner(s).

%%%%%%%%% REFERENCES
% \clearpage
{\small
\bibliographystyle{ieee_fullname}
\bibliography{references}
}

%%%%%%%%% APPENDIX
\clearpage
\onecolumn
\appendix
\section{Additional Ablation Study Results}
\label{sec:additional-ablation-study}

\subsection{Qualitative Results for Ablation Studies}

\begin{figure}[!h]
  \centering
  \resizebox{\linewidth}{!}{
    \includegraphics{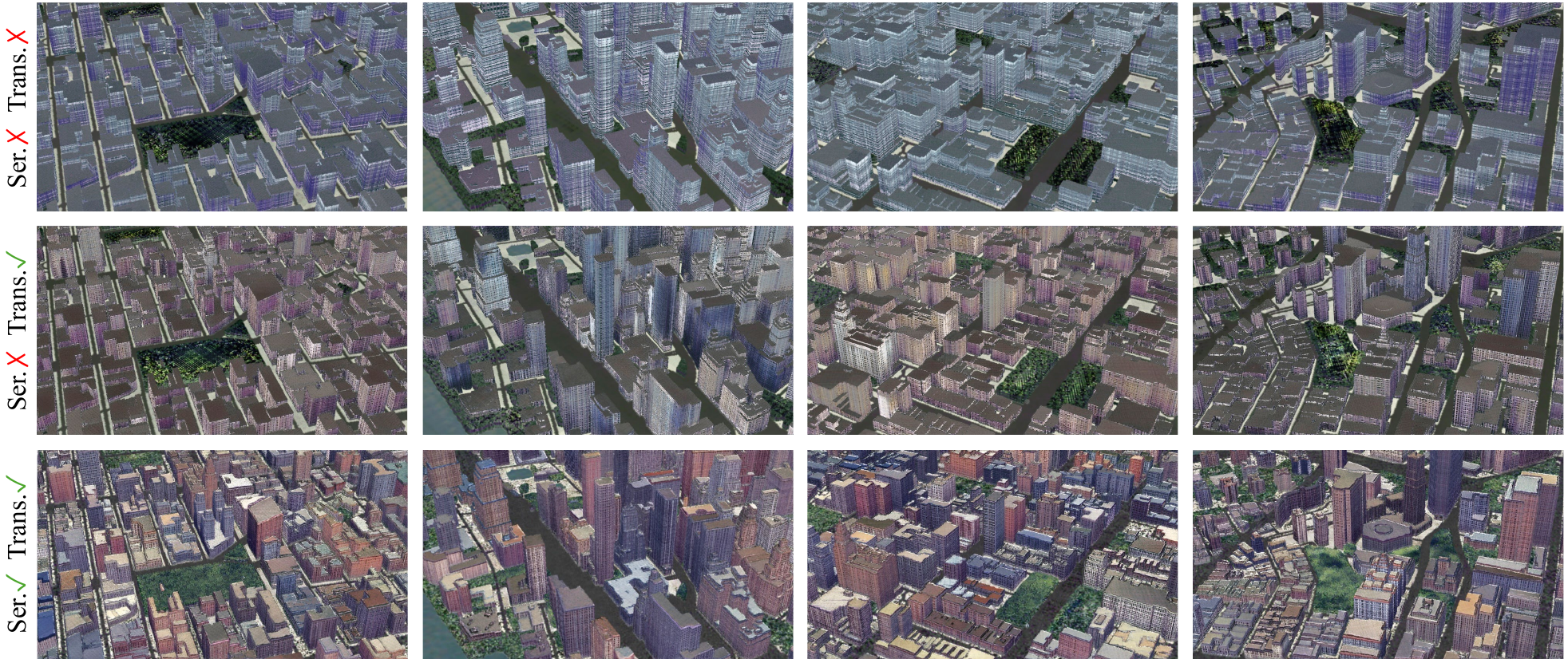}
  }
  \caption{\textbf{Qualitative comparison of different BEV-Point Decoder variants.} Note that ``Ser.'' and ``Trans.'' represent ``Point Serializer'' and ``Point Transformer'', respectively.}
  \label{fig:ablation-bev-point-decoder}
\end{figure}

\noindent \textbf{Effectiveness of BEV-Point Decoder.}
Figure~\ref{fig:ablation-bev-point-decoder} gives a qualitative comparison as a supplement to Table~\ref{tab:ablation-bev-point-decoder}, demonstrating the effectiveness of Point Serializer and Point Transformer in BEV-Point Decoder.
Removing either of them significantly degrades the quality of the generated images.

\begin{figure}[!h]
  \centering
  \resizebox{\linewidth}{!}{
    \includegraphics{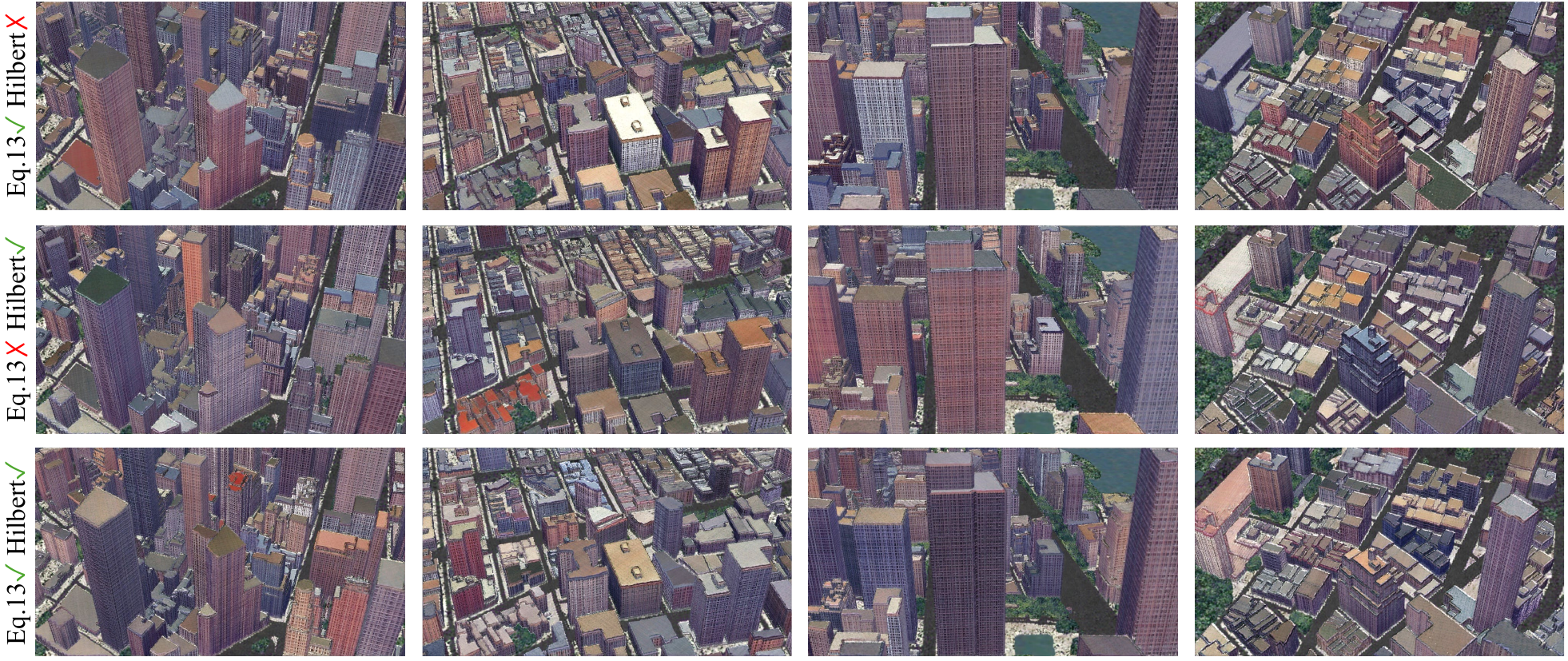}
  }
 \caption{\textbf{Qualitative comparison of different serialization methods.} Note that ``Eq.~\ref{eq:serialization}'' denote the serialization used in Point Serializer.}
  \label{fig:ablation-serialization}
\end{figure}

\noindent \textbf{Effectiveness of Different Serialization Methods.}
Figure~\ref{fig:ablation-serialization} provides a qualitative comparison as a supplement to Table~\ref{tab:ablation-serialization}.
As shown in Figure~\ref{fig:ablation-serialization}, using either Equation~\ref{eq:serialization}, the Hilbert curve, or both for serialization is unlikely to impact the quality of the generated results. 
Moreover, Equation~\ref{eq:serialization} exhibits significantly lower computational complexity compared to the Hilbert curve. 
Therefore, Equation~\ref{eq:serialization} is employed in the Point Serializer due to its lower computational demands.

\clearpage
\subsection{Discussion on Generated Gaussian Attributes}
\label{sec:gaussian-attributes}

In 3D-GS~\cite{DBLP:journals/tog/KerblKLD23}, each 3D Gaussian possesses multiple attributes, including XYZ coordinates, spherical harmonics (SHs), opacity, rotation, and scale. 
These attributes are optimized using supervision from multi-view images to represent the scene.
In reconstruction, adding extra attributes like XYZ offsets and opacity enhances representation capability. 
However, a scene can have multiple valid reconstruction results because different combinations of 3D Gaussian attributes can yield the same rendering result. 
For instance, changing the color of one 3D Gaussian can be equivalent to overlaying multiple Gaussians with varying opacities or adjusting the scale of nearby Gaussians with similar colors. 
This ambiguity causes instability when optimizing all these attributes simultaneously in city generation.

\begin{figure}[!h]
  \centering
  \vspace{-5 mm}
  \resizebox{\linewidth}{!}{
    \includegraphics{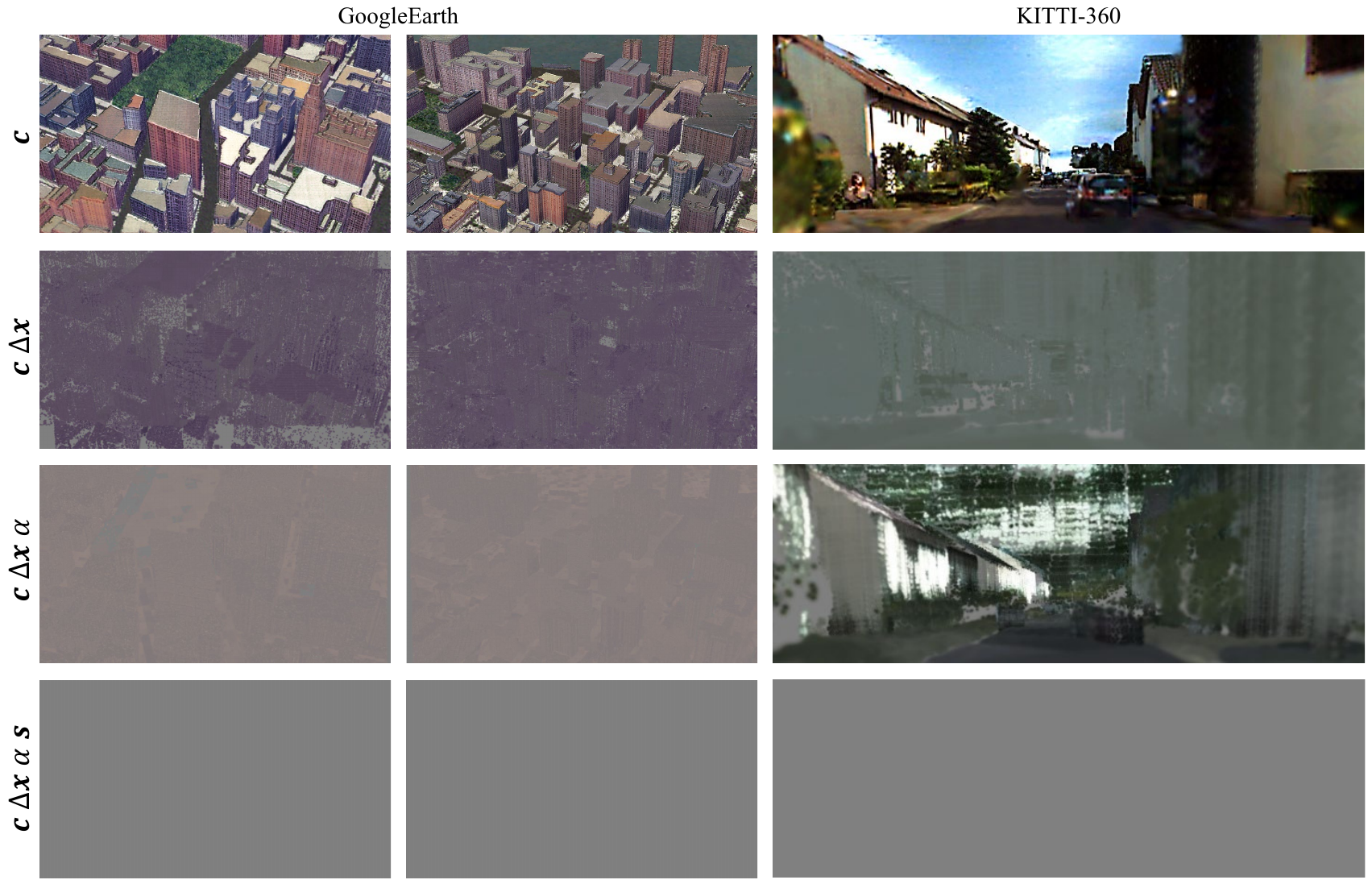}
  }
  \vspace{-5 mm}
  \caption{\textbf{Qualitative comparison of different attributes generated for 3D Gaussians.} Note that $\mathbf{c}$, $\Delta \mathbf{x}$, $\alpha$, and $\mathbf{s}$ denote the RGB color, XYZ offsets, opacity, and scale in the generated 3D Gaussian attributes, respectively.}
  \label{fig:ablation-gaussian-attr}
  \vspace{-5 mm}
\end{figure}

Due to the carefully designed BEV-Point initialization, all points are evenly distributed on the surface of objects. 
Therefore, Gaussian attributes other than RGB can be left unestimated and set to their default values.
Table~\ref{tab:ablation-gaussian-attr} and Figure~\ref{fig:ablation-gaussian-attr} provide a quantitative and qualitative comparison of various attributes generated for 3D Gaussians, demonstrating that optimizing 3D Gaussian attributes beyond RGB not only complicates network convergence but also significantly impacts the quality of the generated results.

\begin{table}[!h]
  \setlength\extrarowheight{1pt}
  \caption{\textbf{Quantitative comparison  of different attributes generated for 3D Gaussians.} The best results are highlighted in bold. Note that ``-'' in CE indicates that COLMAP cannot estimate camera poses from the generated images.}
  \vspace{-2 mm}
  \label{tab:ablation-gaussian-attr}
  \begin{tabularx}{\linewidth}{YYYY|YYYY|YY}
    \toprule
      \multicolumn{4}{c|}{Generated Attributes} & 
      \multicolumn{4}{c|}{GoogleEarth} & 
      \multicolumn{2}{c}{KITTI-360} \\
      \cline{1-4} \cline{5-8} \cline{9-10}
        RGB              & $\Delta$XYZ
      & Opacity          & Scale
      & FID $\downarrow$ & KID $\downarrow$ 
      & DE $\downarrow$  & CE $\downarrow$ 
      & FID $\downarrow$ & KID $\downarrow$ \\
    \midrule
      \cmark     & \xmark     & \xmark     & \xmark     &
      \bf{86.94} & \bf{0.090} & \bf{0.136} & \bf{0.057} & 
      \bf{29.5}  & \bf{0.017} \\
      \cmark     & \cmark     & \xmark     & \xmark     &
      371.15     & 0.468      & 0.367      & -          & 
      281.5      & 0.342 \\
      \cmark     & \cmark     & \cmark     & \xmark     &
      384.81     & 0.485      & 0.401      & -          & 
      256.9      & 0.297 \\
      \cmark     & \cmark     & \cmark     & \cmark     &
      535.49     & 0.709      & 0.470      & -          & 
      486.3      & 0.605 \\
    \bottomrule
  \end{tabularx}
\end{table}

\clearpage
\section{Additional Experimental Results}
\label{sec:additional-experiments}

% \subsection{View Consistency Comparison}

% \begin{figure}[!h]
%   \centering
%   \resizebox{\linewidth}{!}{
%     \includegraphics{figures/colmap-reconstruction}
%   }
%  \caption{Caption}
%   \label{fig:colmap-reconstruction}
% \end{figure}

\subsection{Building Interpolation}

\begin{figure}[!h]
  \centering
  \resizebox{\linewidth}{!}{
    \includegraphics{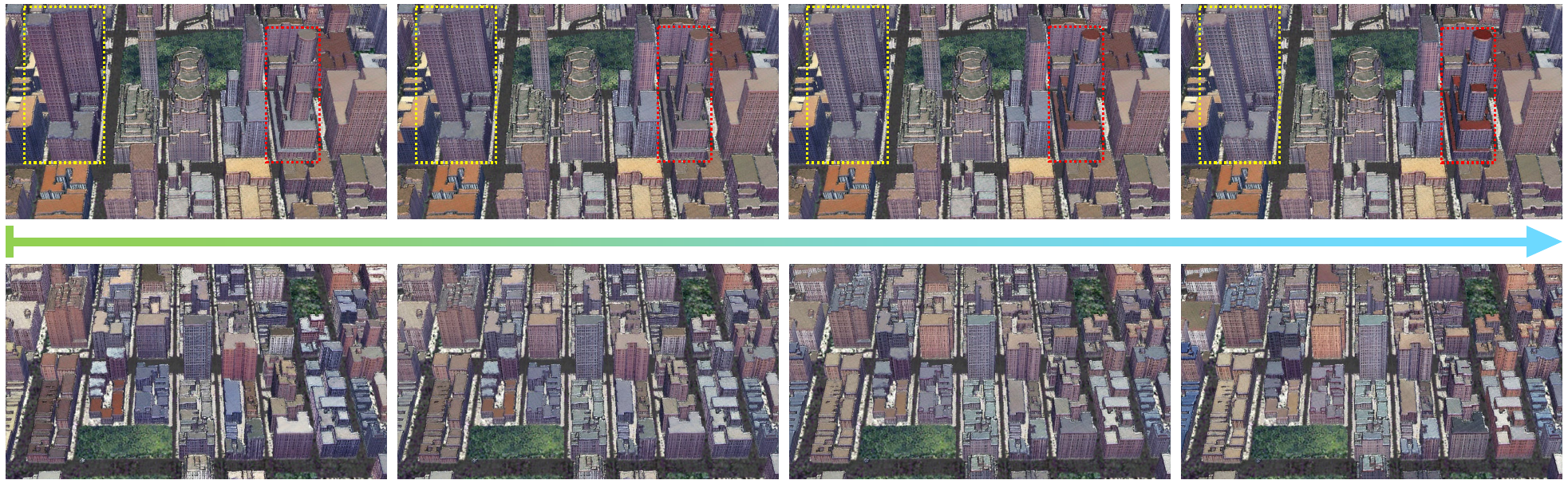}
  }
 \caption{\textbf{Linear interpolation along the building style.} The style of each building gradually changes from left to right. In the first row, only the styles of the buildings enclosed by the bounding boxes are modified, while in the second row, the styles of all the buildings are changed.}
  \label{fig:bldg-interpolation}
\end{figure}

As shown in Figure~\ref{fig:bldg-interpolation}, GaussianCity showcases the capability to interpolate building styles controlled by the variable $z$.
In the first row, local editing is applied by altering only the style code $z$ of the buildings within the bounding boxes, leaving the styles of the other buildings unchanged. 
In the second row, the style codes of all the buildings are interpolated from left to right, resulting in a gradual style transition across the entire row.

% \subsection{Localized Editing}

% Each building instance is uniquely controlled by a style code $z$ in GaussianCity, enabling localized editing of building instances. 
% Figure~\ref{fig:local-editing} demonstrates how the style and height of each building instance can be independently adjusted.

% \begin{figure}[!h]
%   \centering
%   \resizebox{\linewidth}{!}{
%     \includegraphics{figures/local-editing}
%   }
%   \caption{\textbf{Localized editing for the building instances highlighted with bounding boxes.} (a) Moving from left to right, the style of several buildings changes gradually, while the rest remain unchanged. (b) The building's style remains constant while its appearance adjusts dynamically to varying heights. (c) The styles of the two buildings can be interchanged. (d) Applying a new style vector changes the building's appearance.}
%   \label{fig:local-editing}
% \end{figure}

\subsection{Relighting}

\begin{figure}[!h]
  \centering
  \resizebox{\linewidth}{!}{
    \includegraphics{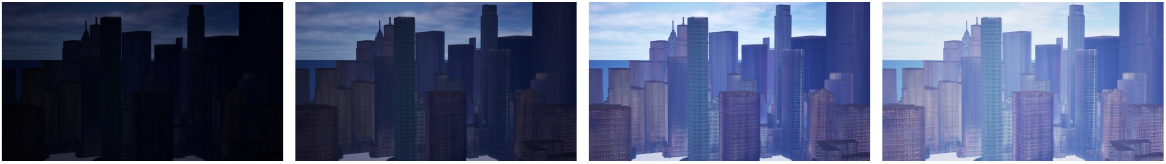}
  }
  \caption{\textbf{Relighting effects in Unreal Engine 5.} From left to right, the relighting effect is shown with increasing light intensity.}
  \label{fig:relighting}
\end{figure}

From the explicit representation of 3D Gaussians, relighting is much simpler than NeRF-based methods. 
Using Luma AI's 3D Gaussians Plugin~\footnote{\url{https://www.unrealengine.com/marketplace/product/66dd775fa3104ecfb3ae800b8963c8b9}}, the generated city can be imported into Unreal Engine, enabling highly realistic lighting effects, as illustrated in Figure~\ref{fig:relighting}.
However, the plugin is still under development, so shadow effects are not well generated. 
We believe this issue will be resolved soon.

\clearpage
\subsection{Additional Qualitative Comparison}

\begin{figure}[!h]
  \centering
  \resizebox{\linewidth}{!}{
    \includegraphics{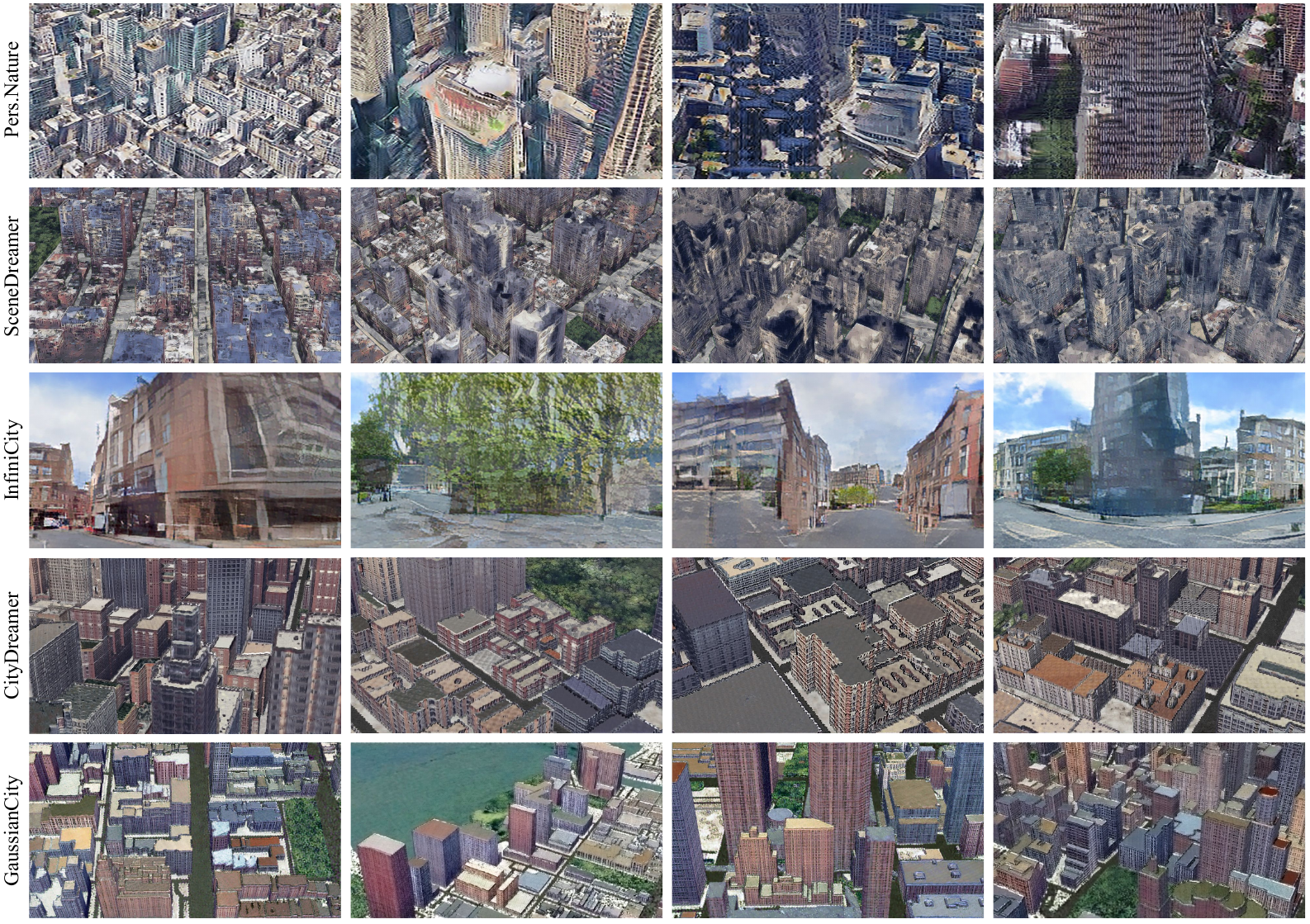}
  }
  \caption{\textbf{Qualitative comparison on GoogleEarth.} Note that ``Pers.Nature'' is short for ``PersistentNature''~\cite{DBLP:conf/cvpr/Chai0LIS23}. The visual results of InfiniCity~\cite{DBLP:conf/iccv/LinLMCS0T23} are provided by the authors and zoomed for optimal viewing.}
  \label{fig:qualitative-google-earth-extra1}
\end{figure}

\begin{figure}[!h]
  \centering
  \resizebox{\linewidth}{!}{
    \includegraphics{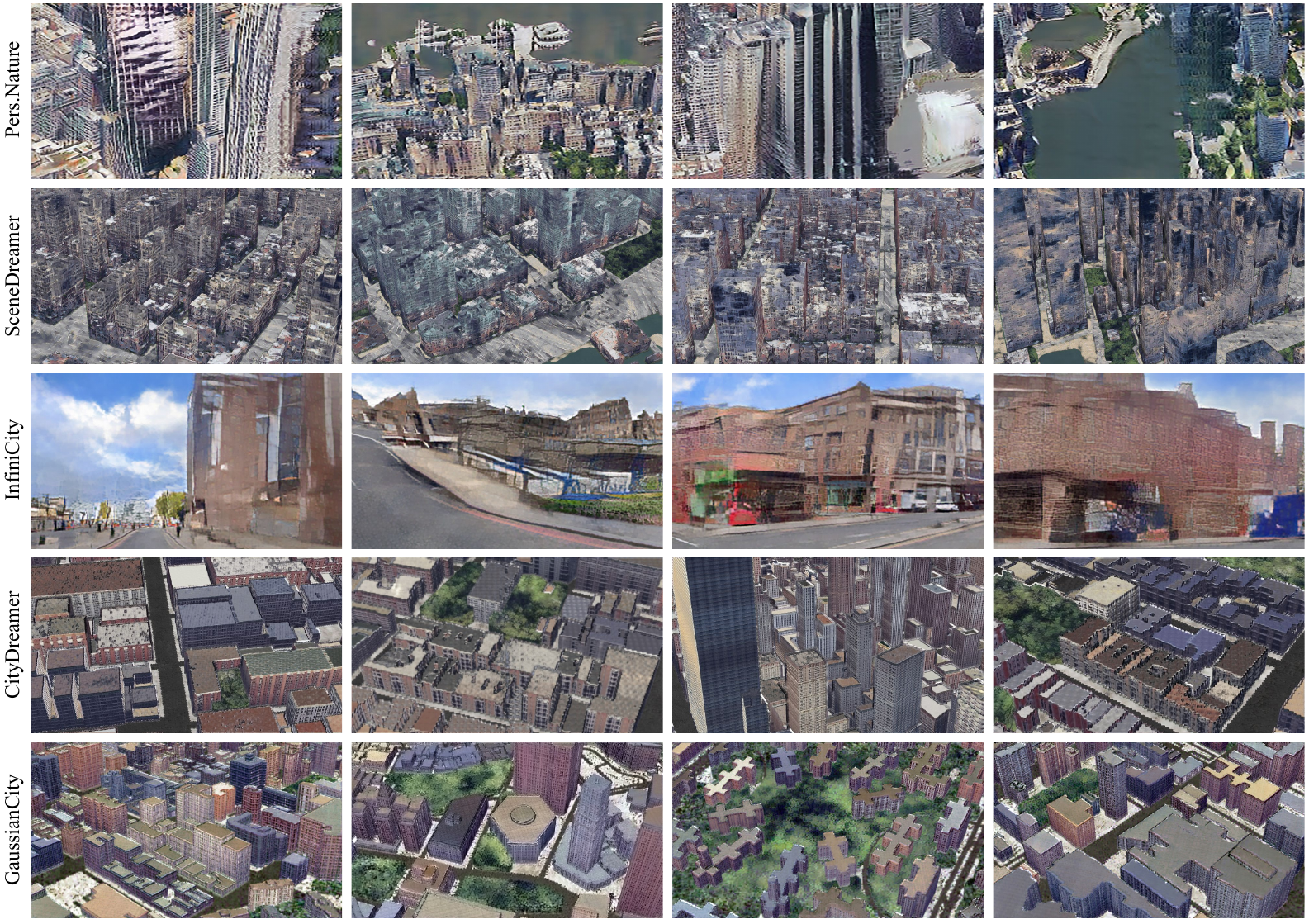}
  }
  \caption{\textbf{Qualitative comparison on GoogleEarth.} Note that ``Pers.Nature'' is short for ``PersistentNature''~\cite{DBLP:conf/cvpr/Chai0LIS23}. The visual results of InfiniCity~\cite{DBLP:conf/iccv/LinLMCS0T23} are provided by the authors and zoomed for optimal viewing.}
  \label{fig:qualitative-google-earth-extra2}
\end{figure}

\begin{figure}[!t]
  \centering
  \resizebox{\linewidth}{!}{
    \includegraphics{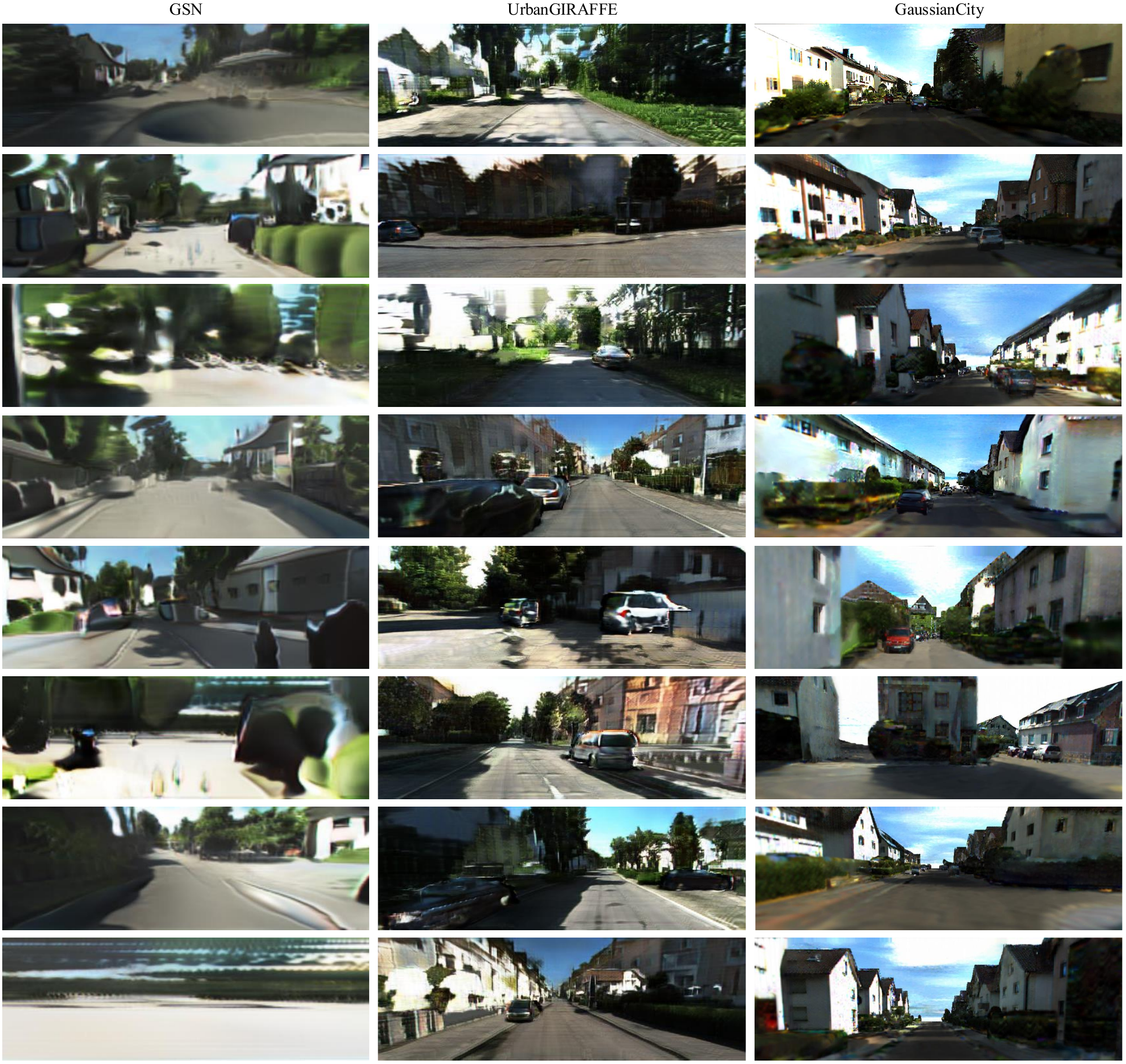}
  }
  \caption{\textbf{Qualitative comparison on KITTI-360.} The visual results of UrbanGIRAFFE~\cite{DBLP:conf/iccv/YangYGX0L23} are provided by the authors since the training code and pretrained model are unavailable.}
  \label{fig:qualitative-kitti360-extra}
\end{figure}

In Figures~\ref{fig:qualitative-google-earth-extra1},~\ref{fig:qualitative-google-earth-extra2} and~\ref{fig:qualitative-kitti360-extra}, we provide more visual comparisons with state-of-the-art methods on GoogleEarth and KITTI-360, respectively.

\clearpage
\null
\vfill

\clearpage
\null
\vfill

% \clearpage
% \section{Licenses and URLs for Used Assets}
% \label{sec:licenses}

% \subsection{Datasets}

% \noindent \textbf{GoogleEarth}~\cite{DBLP:conf/cvpr/XieCHL24}.
% %
% S-Lab License.
% %
% \url{https://paperswithcode.com/dataset/googleearth}

% \noindent \textbf{KITTI-360}~\cite{DBLP:journals/pami/LiaoXG23}. 
% %
% CC BY-NC-SA 3.0.
% %
% \url{https://paperswithcode.com/dataset/kitti-360}

% \subsection{Models}

% \noindent \textbf{PersistentNature}~\cite{DBLP:conf/cvpr/Chai0LIS23}.
% %
% Apache-2.0 License.
% %
% \url{https://github.com/google-research/google-research/tree/master/persistent-nature}

% \noindent \textbf{SceneDreamer}~\cite{DBLP:journals/pami/ChenWL23}.
% %
% S-Lab License.
% %
% \url{https://github.com/FrozenBurning/SceneDreamer}

% \noindent \textbf{CityDreamer}~\cite{DBLP:conf/cvpr/XieCHL24}.
% %
% S-Lab License.
% %
% \url{https://github.com/hzxie/CityDreamer}

% \noindent \textbf{GSN}~\cite{DBLP:conf/iccv/DeVries0STS21}.
% %
% Apple License.
% %
% \url{https://github.com/apple/ml-gsn}

% \noindent \textbf{GIRAFFE}~\cite{DBLP:conf/cvpr/Niemeyer021}.
% %
% MIT License.
% %
% \url{https://github.com/autonomousvision/giraffe}

% \noindent \textbf{MiDaS}~\cite{DBLP:journals/pami/RanftlLHSK22}.
% %
% MIT License.
% %
% \url{https://github.com/isl-org/MiDaS}

% \subsection{Others}
% %
% \noindent \textbf{COLMAP} (Ver. 3.9.1)~\cite{DBLP:conf/cvpr/SchonbergerF16}.
% %
% BSD License.
% %
% \url{https://github.com/colmap/colmap}

% \noindent \textbf{Unreal Engine} (Ver. 5.3.2). Free License. 
% \url{https://www.unrealengine.com/unreal-engine-5}

% \noindent \textbf{Luma AI's UE Plugin} (Ver. 0.41). Free License. 
% \url{https://www.unrealengine.com/marketplace/product/66dd775fa3104ecfb3ae800b8963c8b9}

\end{document}